\documentclass{article}

\usepackage{arxiv}

\usepackage[utf8]{inputenc} 
\usepackage[T1]{fontenc}    
\usepackage{hyperref}       
\usepackage{url}            
\usepackage{booktabs}       
\usepackage{amsfonts}       
\usepackage{nicefrac}       
\usepackage{microtype}      
\usepackage{lipsum}
\usepackage{graphicx}
\graphicspath{ {./images/} }
\usepackage{natbib}
\usepackage{multirow}
\title{Causality-Driven Infrared and Visible Image Fusion}

\author{
Linli Ma \\
School of Computer Science and Technology\\
  North University of China\\
  \texttt{b20230706@st.nuc.edu.cn} \\
   \And
Suzhen Lin \\
School of Computer Science and Technology\\
  North University of China\\
  \texttt{lsz@nuc.edu.cn} \\
  \And
Jianchao Zeng \\
School of Computer Science and Technology\\
  North University of China\\
  \texttt{zjc@nuc.edu.cn} \\
    \And
Zanxia Jin \\
School of Computer Science and Technology\\
  North University of China\\
  \texttt{zanxiajin@nuc.edu.cn} \\
    \And
Yanbo Wang \\
School of Computer Science and Technology\\
  North University of China\\
  \texttt{yanbowang@nuc.edu.cn} \\
    \And
Fengyuan Li \\
School of Computer Science and Technology\\
  North University of China\\
  \texttt{s202207011@st.nuc.edu.cn} \\
    \And
Yubing Luo \\
School of Computer Science and Technology\\
  North University of China\\
  \texttt{sz202107003@st.nuc.edu.cn} \\
}

\begin{document}
\maketitle
\begin{abstract}
Image fusion aims to combine complementary information from multiple source images to generate more comprehensive scene representations. Existing methods primarily rely on the stacking and design of network architectures to enhance the fusion performance, often ignoring the impact of dataset scene bias on model training. This oversight leads the model to learn spurious correlations between specific scenes and fusion weights under conventional likelihood estimation framework, thereby limiting fusion performance. To solve the above problems, this paper first re-examines the image fusion task from the causality perspective, and disentangles the model from the impact of bias by constructing a tailored causal graph to clarify the causalities among the variables in image fusion task. Then, the Back-door Adjustment based Feature Fusion Module (BAFFM) is proposed to eliminate confounder interference and enable the model to learn the true causal effect. Finally, Extensive experiments on three standard datasets prove that the proposed method significantly surpasses state-of-the-art methods in infrared and visible image fusion. 
\end{abstract}


\section{Introduction}

For complex imaging scenarios, single sensors struggle to comprehensively capture critical scene information owing to the constraints of shooting environment or hardware \citep{rf1}. To address this challenge, image fusion, as a key image enhancement technique, properly integrates complementary information from multiple sensors, significantly improving scene perception and understanding \citep{rf2,rf3}. Specifically, visible sensors capture texture details through reflected light but are highly susceptible to environmental elements like light intensity and atmospheric conditions \citep{rf4,rf5}. Conversely, infrared sensors generate images by capturing thermal radiation, enabling high-contrast target representation under all-weather conditions but lacking texture details \citep{rf6}. This inherent spectral complementarity makes infrared and visible image fusion essential for applications like object detection, tracking and security surveillance \citep{rf7,rf8}.

As deep learning advances quickly, deep learning-based image fusion methods have gradually become mainstream \citep{rf9}, which mainly include methods based on auto-encoder (AE) \citep{rf10,rf11}, convolutional neural networks (CNN) \citep{rf12,rf13}, generative adversarial networks (GAN) \citep{rf14,rf15}, and Transformer \citep{rf16,rf17}. These methods construct a nonlinear mapping from source to fused images via end-to-end learning, achieving significant progress in feature representation and fusion quality. However, due to the lack of labels as reference for image fusion task, existing methods primarily rely on loss constraints between fused and source images for optimization \citep{rf18}. This makes the model susceptible to dataset scene bias during training, leading it to learn spurious correlations between specific scenes and fusion weights, thereby severely limiting fusion performance. Specifically, deep learning models update their parameters by backpropagating gradients to minimize the loss constraints \citep{rf19}.  When dataset’s scene distribution is imbalanced, high-frequency scenes tend to dominate the loss calculation, causing the model to form a stable gradient update pattern in high-frequency scenes, while for low-frequency scenes, the gradient update is relatively insufficient. The imbalance of this gradient optimization can lead to the model tending to learn fusion strategies in high-frequency scenes during training (i.e. establishing spurious correlations between high-frequency scenes and fusion weights), rather than accurately capturing the true cross-modal information complementarity. Therefore, during the testing phase, when the model processes scenes that have not been seen or have a low proportion in the training set, its fusion strategy will deviate from the optimal solution, inevitably leading to artifacts or information loss in the fusion results. 

\begin{figure}
	\centering
	\includegraphics[scale=0.2]{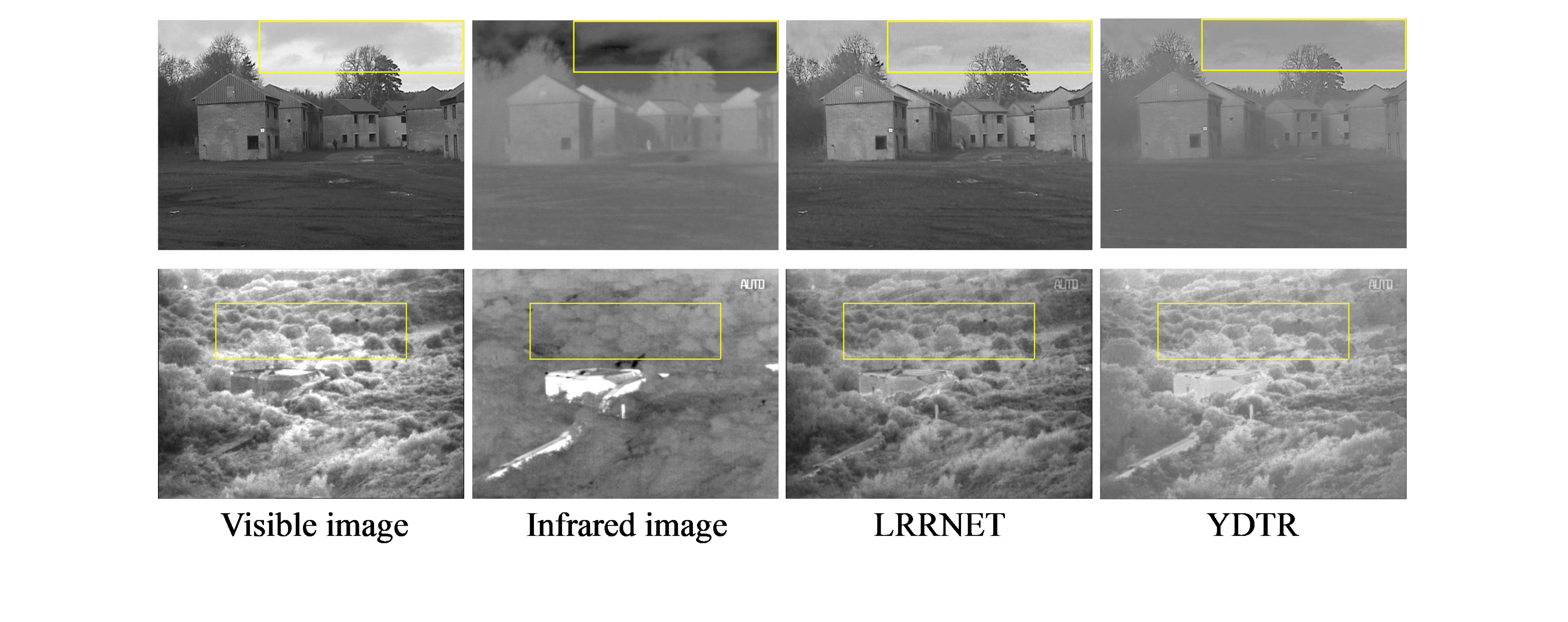}
	\caption{Fusion bias problem in natural scenes. The training sets of LRRNET and YDTR models are mainly based on street scenes. When predicting natural scenes such as cloud and bush, the fusion results may deviate, leading to artifacts and information loss.}
	\label{FIG:1}
\end{figure}

The research results of LRRNET \citep{rf13} and YDTR \citep{rf20} have confirmed the above phenomenon. Specifically, the training sets of these two studies mainly focus on street scenes, leading the models to implicitly learn spurious correlations between street scenes and fusion weights during training. When testing on the TNO dataset, which includes a multiple real-world scenes such as urban, natural landscapes, and military surveillance zones \citep{rf21}, artifacts and information loss occurred in the fusion results. As shown in Fig.~\ref{FIG:1}, thermal radiation noise artifacts were generated in the cloud area, while critical information was lost in bush-covered areas. This indicates that the model has developed a bias toward street scenes, making it incapable of accurately estimating fusion weights when processing low-frequency scenes (e.g., clouds and bush), thereby compromising the overall fusion quality.

Given the above observations, this paper proposes a novel causality-driven infrared and visible image fusion method. Specifically, dataset scene bias can mislead the model into learning spurious correlations between specific scenes and fusion weights. Therefore, we treat dataset scene bias as a confounder and construct tailored causal graph to systematically describe the causalities among variables in image fusion. Building on this foundation, we propose a Back-door Adjustment based Feature Fusion Module (BAFFM), which leverages do-calculus from causal inference to estimate the true causal effect, thereby eliminating the confounder’s impact on the fusion results. The key contributions of this paper are summarized below:

\begin{itemize} \item This study is the first to explore image fusion tasks through causal inference, proposing a new causality-driven infrared and visible image fusion method, effectively alleviating the problem of unreasonable fusion weight prediction caused by dataset scene bias.
                \item A tailored causal graph is constructed to systematically describe the causalities among variables in the image fusion task, providing a new theoretical framework for image fusion.  
                \item A Back-door Adjustment based Feature Fusion Module (BAFFM) is proposed to ensure that different scenes can contribute more equitably to the model training, thereby improving fusion performance.
                 \item Extensive experiments on three standard infrared and visible image fusion datasets demonstrate that the proposed method outperforms state-of-the-art methods.
\end{itemize}

\section{Related Work}
\subsection{Image Fusion}

Recently, deep learning has emerged as the dominant technology in image fusion thanks to its powerful feature representation ability \citep{rf22}. Depending on the network architectures, deep learning-based image fusion methods can be categorized into four categories: 1) AE-based methods: Encoder-decoder networks are typically trained using general datasets like MS-COCO \citep{rf23}. The core process involves: the encoder extracts deep features from source images→fuses features based on heuristic rules (e.g., L1-norm \citep{rf24}, dynamic weight selection \citep{rf25}) → the decoder reconstructs the fused result \citep{rf26}. 2) CNN-based methods: This method inherits the core idea of optimization-based and implicitly performs feature extraction, fusion, and reconstruction through complex loss functions, thereby improving the problem of manually designing fusion rules \citep{rf27,rf28,rf29}. 3) GAN-based methods: This method utilizes the interaction of the generator and discriminator to implicitly learn the data distribution characteristics. Reference \citep{rf30} was the first to apply GAN to image fusion, encouraging the generator to produce fused images with rich texture details using the discriminator’s ability. Nevertheless, a single discriminator may result in modal collapse issues. Therefore, \citep{rf31,rf32,rf33} further proposed a hierarchical discrimination strategy to balance the retention strength of multi-modal information. 4) Transformer-based methods: With the long-range dependency modeling capability of self-attention mechanism, this method uses Transformer structures to replace or combine with traditional CNN structures \citep{rf34,rf35}, providing a new architectural guidelines for image fusion. This enhances the capability to capture global information, overcoming the limitations of  CNNs' local receptive fields \citep{rf36}. Although the above methods have achieved significant performance improvements through data-driven optimization, they generally overlook the impact of dataset scene bias on the model training. Therefore, how to alleviate the impact of scene bias is still a crucial challenge in the field of image fusion.

\subsection{Causal Inference}

Causal inference, as an useful tool, has gradually been widely applied to solve the problem of dataset bias \citep{rf37,rf38}. Existing causal inference methods primarily control confounders through front-door or back-door adjustment strategies, ensuring that models focus on true causal effects without being influenced by potential external biases \citep{rf39}. For instance, \citep{rf40} proposed a Contextual Causal Intervention Module (CCIM) based on back-door adjustment, aiming to eliminate context bias and promote fair contributions from different contexts in emotion understanding. \citep{rf41} introduced a causal inference based image captioning (CIIC) framework, which significantly improved image captioning accuracy by eliminating the interference of visual and linguistic confounders in datasets. Additionally, \citep{rf42} developed causal learning modules based on both back-door and front-door adjustments (BACL and FACL), which enhance unbiased learning by comprehensively reducing potential spurious correlations. Due to the observability of dataset scene categories and their distribution frequencies, confounder can be identified and analyzed. Therefore, we choose the back-door adjustment method to address the problem of unreasonable fusion weight prediction caused by dataset scene bias.

\section{Methodology}
\label{sec3}
\subsection{Causal View}
\label{subsec31}
In image fusion, dataset scene bias can interfere with model's learning process, leading the allocation of fusion weights to deviate from the optimal solution, thereby affecting the final fusion result. To systematically characterize this phenomenon and the causalities among various variables, we conducted an in-depth analysis of the fusion process and gradually derived a tailored causal graph.

First, due to the constraints of physical conditions, the scenes coverage of image fusion datasets is limited. Because of this, specific scenes occupy a large proportion of the training set, while others are relatively small or even missing. We refer to this imbalanced scene distribution as dataset scene bias, which is treated as a confounder $Z$ in causal inference. The image features $X$ and $Z$ jointly determine the biased content of the fusion weights, i.e., $Z \rightarrow X$.

Secondly, since the model's loss calculation is primarily influenced by high-frequency scenes in the training set, the optimization process tends to learn spurious correlations between high-frequency scenes and fusion weights rather than capturing true cross-modal complementary information. This implies that the confounder $Z$ leads the model to learn unreliable fusion weights W, i.e., $Z \rightarrow W$. Furthermore, the fusion weights $W$ directly determine the final fused image $Y$, i.e., $W \rightarrow Y$. Therefore, the causal path $Z \rightarrow W \rightarrow Y$ illustrates that dataset scene bias affects the final fused image $Y$ by influencing the fusion weights $W$.

Finally, in the image fusion process, the model calculates the fusion weights $W$ relying on the input infrared and visible image features $X$ to determine the contribution of different modalities in the fused image, leading to the fused image $Y$, i.e., $X \rightarrow W \rightarrow Y$. Additionally, image features $X$ itself determines the fundamental information of $Y$. Whatever how the $W$ are adjusted, the structure, texture, and intensity features of $X$ will be selectively transmitted to $Y$, i.e., $X \rightarrow Y$. Therefore, the impact of $X$ on the $Y$ can be summarized into two primary paths: $X \rightarrow W \rightarrow Y$ and $X \rightarrow Y$.

\begin{figure}
	\centering
	\includegraphics[scale=1.3]{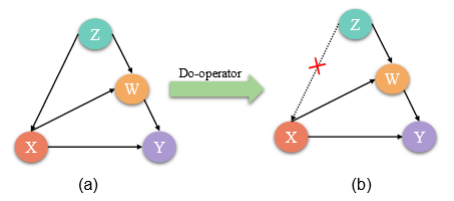}
	\caption{Causal graph of image fusion (a) Conventional likelihood $P\left(Y|X\right)$ (b) Causal intervention $P\left(Y|do\left(X\right)\right)$.}
	\label{FIG:2}
\end{figure}

In summary, the causal graph for image fusion is illustrated in Fig.~\ref{FIG:2}(a), involving four variables: image features $X$, fusion weights $W$, fused image $Y$, and confounder $Z$. Ideally, the model should follow the causal paths $X \rightarrow W \rightarrow Y$ and $X \rightarrow Y$ to ensure that it learns true cross-modal complementary information. However, based on causal theory \citep{rf43}, $Z$ can mislead model into learning spurious correlations through the back-door causal path $X \leftarrow Z \rightarrow W \rightarrow Y$, thereby interfering with the final fusion result.
\subsection{Causal Intervention}
\label{subsec32}
According to Fig.~\ref{FIG:2}(a), existing fusion methods use conventional likelihood estimation $P\left(Y|X\right)$ to predict the fusion result, which is inevitably influenced by the confounder $Z$. Based on Bayes rule, this process is formulated as:
\begin{equation}\label{eq1}
     P\left(Y|X\right)= \sum_{z} P\left(Y|X,W=f_{c}\left(X,z\right)\right)P\left(z|X\right)
\end{equation}
where $f_{c}\left( \cdot \right)$ is a generalized encoding function.

To tackle the adverse confounding effects introduced by $Z$ and ensure that fusion model estimates fusion weights solely based on image features $X$, a straightforward approach is to apply an intervention on $X$, forcing different scenes to contribute fairly to the prediction of fusion weights. However, considering the quantity of scene categories in reality is infinite, directly implementing such an intervention is practically infeasible. To address this issue, we stratify $Z$ using back-door adjustment \citep{rf39} to realize causal intervention $P\left(Y|do\left(X\right)\right)$ and effectively cut off the back-door path between $X$ and $Y$. As depicted in Fig.~\ref{FIG:2}(b), the scene intervention based on back-door adjustment can be expressed as:
\begin{equation}\label{eq2}
    P\left(Y|do\left(X\right)\right)= \sum_{z} P\left(Y|X,W=f_{c}\left(X,z\right)\right)P\left(z\right)
\end{equation}
where $P\left(Y|do\left(X\right)\right)$ aims to ensure that when predicting $Y$, $z$ is not influenced by $X$ and the influence of each $z$ is fairly incorporated, thereby enabling the model to capture the true causalities between $X$ and $Y$.

\begin{figure}
	\centering
	\includegraphics[scale=1.2]{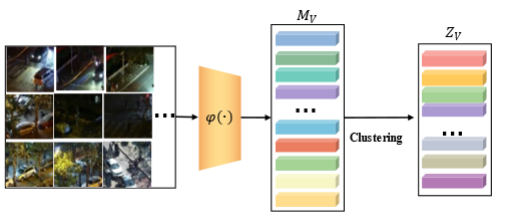}
	\caption{Generation process of visible confounder dictionary $Z_V$.}
	\label{FIG:3}
\end{figure}

Due to the significant differences in the physical imaging principles and the types of information captured by infrared and visible sensors \citep{rf44}, they also exhibit different characteristics in terms of scene bias. To achieve the stratification of $Z$, it is necessary to construct independent confounder dictionaries for each modality. Taking the visible modality as an example, let the visible confounder dictionary be $Z_V=[z_1,z_2,\ldots,z_N]$, where $N$ is a hyperparameter representing the size of dictionary. As depicted in Fig.~\ref{FIG:3}, we initially utilize a pre-trained backbone network $\varphi \left( \cdot \right)$ \citep{rf45} to extract visible scene features from the dataset, forming a visible scene feature set $M_V= \{m_k\}_{k=1}^{N_m}$, where $N_m$ denotes the quantity of training samples.  Then, we apply K-Means++ and Principal Component Analysis (PCA) to obtain $Z_V$, ensuring that each $z_i$ corresponds to the cluster center of a visible scene in the dataset. Each $z_i$ is defined as the mean of all features within its respective cluster, i.e.,  $z_i=\frac{1}{N_i}\sum_{j=1}^{N_i}m_{j}^{i}$, where $N_i$ denotes the quantity of scene features in the $i$-th cluster. Similarly, the process of constructing infrared confounder dictionary $Z_I$ is analogous to the aforementioned visible modality.

\subsection{The Architecture of BAFFM}
\label{subsec3}

To execute the theoretical and hypothetical intervention in eq.~\ref{eq2}, we construct a Back-door Adjustment based Feature Fusion Module (BAFFM) to achieve scene-deconfounded training for the model. Specifically, calculating $P\left(Y|do\left(X\right)\right)$ demands multiple forward passes for all $z$, which incurs high computational costs. Consequently, we use the Normalized Weighted Geometric Mean (NWGM) \citep{rf46} to approximate the computation, expressed as follows:
\begin{equation}\label{eq3}
    P\left(Y|do\left(X\right)\right) \approx P\left(Y|X,W=\sum_{z} f_{c}\left(X,z\right)P\left(z\right)\right)
\end{equation}

We parameterize the above equation into a network model to approximate conditional probability \citep{rf47}:

\begin{equation}\label{eq4}
    P\left(Y|do\left(X\right)\right)= W_{h}X+W_{g}\mathbb{E}_{z}\lbrack{g\left(z\right)}\rbrack
\end{equation}
where $W_h$ and $W_g$ are learnable parameters. $\mathbb{E}_{z}\lbrack{g\left(z\right)}\rbrack$ is approximated as the weighted integration of all scenes:
\begin{equation}\label{eq5}
  \mathbb{E}_{z}\lbrack{g\left(z\right)}\rbrack=\sum_{i=1}^{N}\lambda_{i}z_iP\left(z_i\right)
\end{equation}
where, to ensure the fair inclusion of different scenes, $P\left(z_i\right)=\frac{1}{N}$. $\lambda_{i}$ is a weight coefficient that reflects the importance of each $z_i$ after interacting with $X$:
\begin{equation}\label{eq6}
  \lambda_{i}=softmax\left(\frac{\left(W_{q}X\right)^\mathrm{T}\left(W_{k}z_{i}\right)}{\sqrt{d}}\right)
\end{equation}
where $W_q$ and $W_k$ are learnable parameters.

\begin{figure}
	\centering
	\includegraphics[scale=1.2]{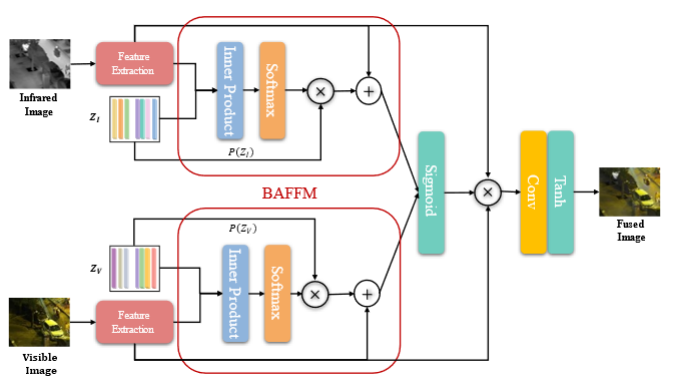}
	\caption{Overall architecture of image fusion network, where the red box represents our proposed BAFFM.}
	\label{FIG:4}
\end{figure}

In summary, our proposed image fusion network architecture is shown in Fig.~\ref{FIG:4}. First, we use the feature extraction module in LRRNET \citep{rf13} to extract image features $X$ from input infrared and visible images. Then, BAFFM applies a back-door adjustment strategy to achieve scene deconfounded training, enabling more precise feature fusion. Specifically, following eq.~\ref{eq3}-~\ref{eq6}, the module computes the interaction importance between input features $X$ and each scene feature $z_i$ in the confounder dictionary using dot product attention, and then approximates the conditional probability through weighted integration to adjust the fusion weights. Finally, the fused features pass through a reconstruction module consisting of a convolutional layer and a Tanh activation function to generate final fused image.
\section{Experimental Results}
\label{sec4}
In this section, we begin by discussing the experimental details, followed by qualitative and quantitative comparisons with state-of-the-art methods. Finally, to validate the effectiveness of proposed method, we conduct ablation studies.
\begin{figure}
	\centering
	\includegraphics[scale=0.2]{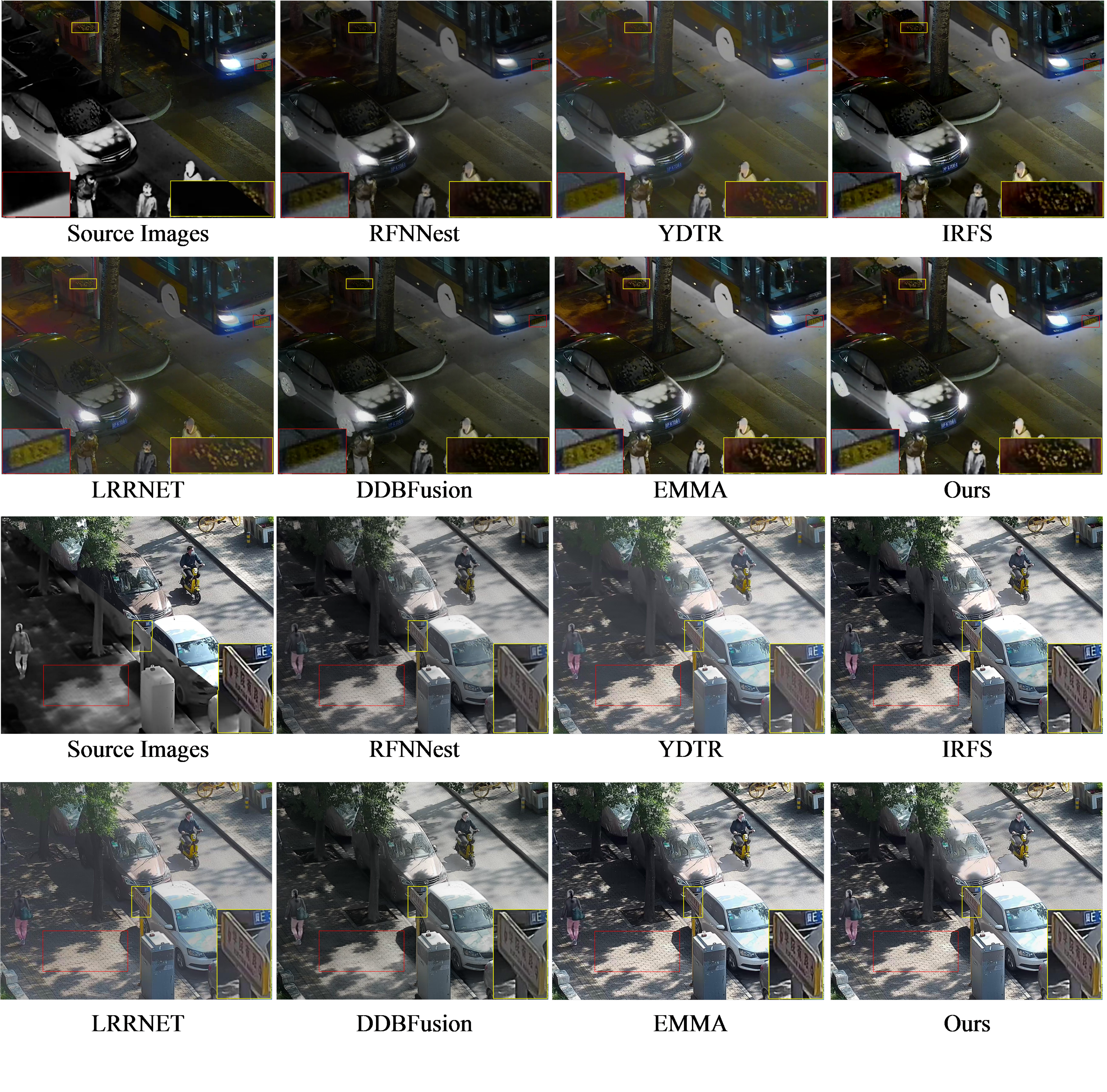}
	\caption{Qualitative comparison on the LLVIP dataset.}
	\label{FIG:5}
\end{figure}

\begin{figure}
	\centering
	\includegraphics[scale=0.2]{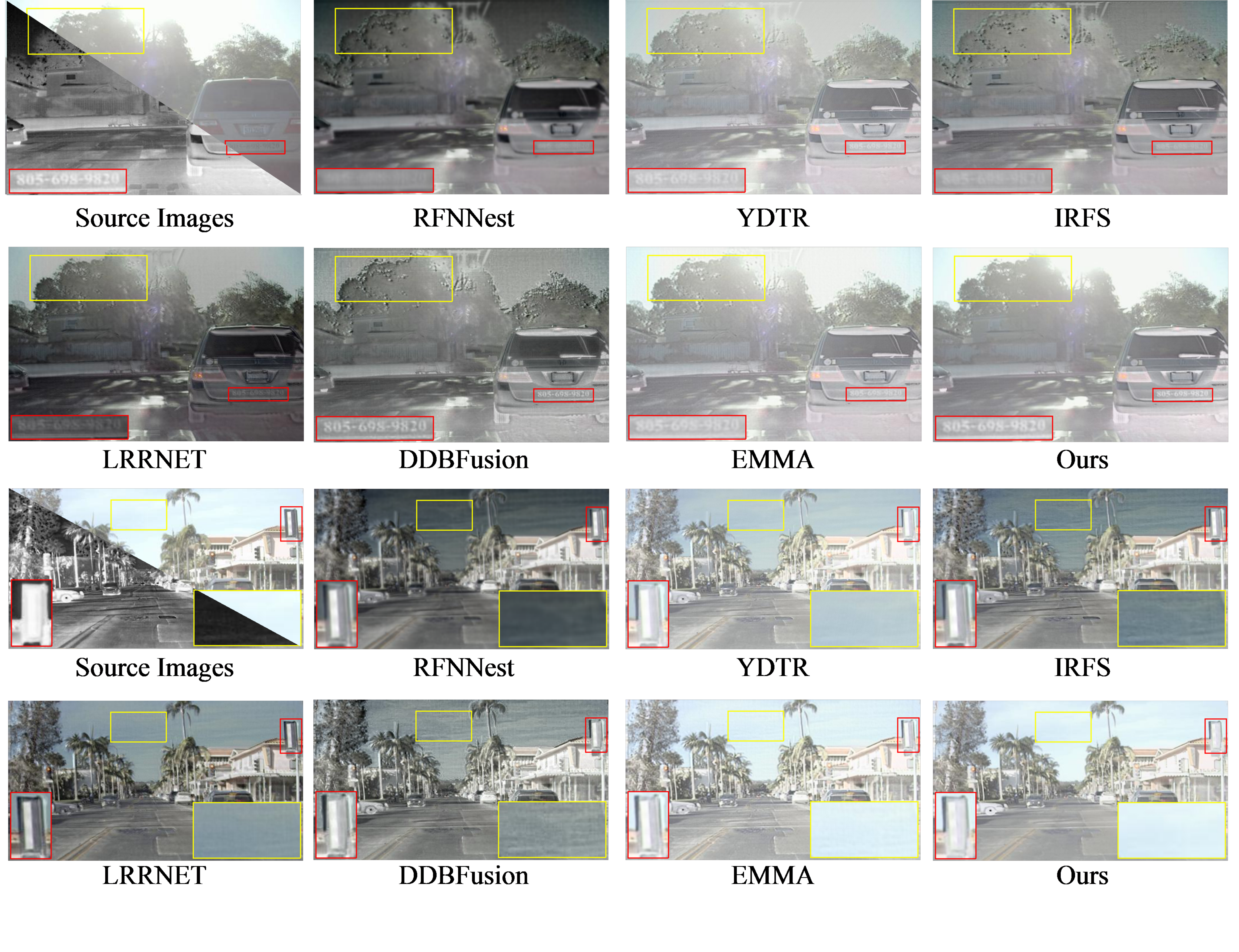}
	\caption{Qualitative comparison on the RoadScene dataset.}
	\label{FIG:6}
\end{figure}

\begin{figure}
	\centering
	\includegraphics[scale=0.18]{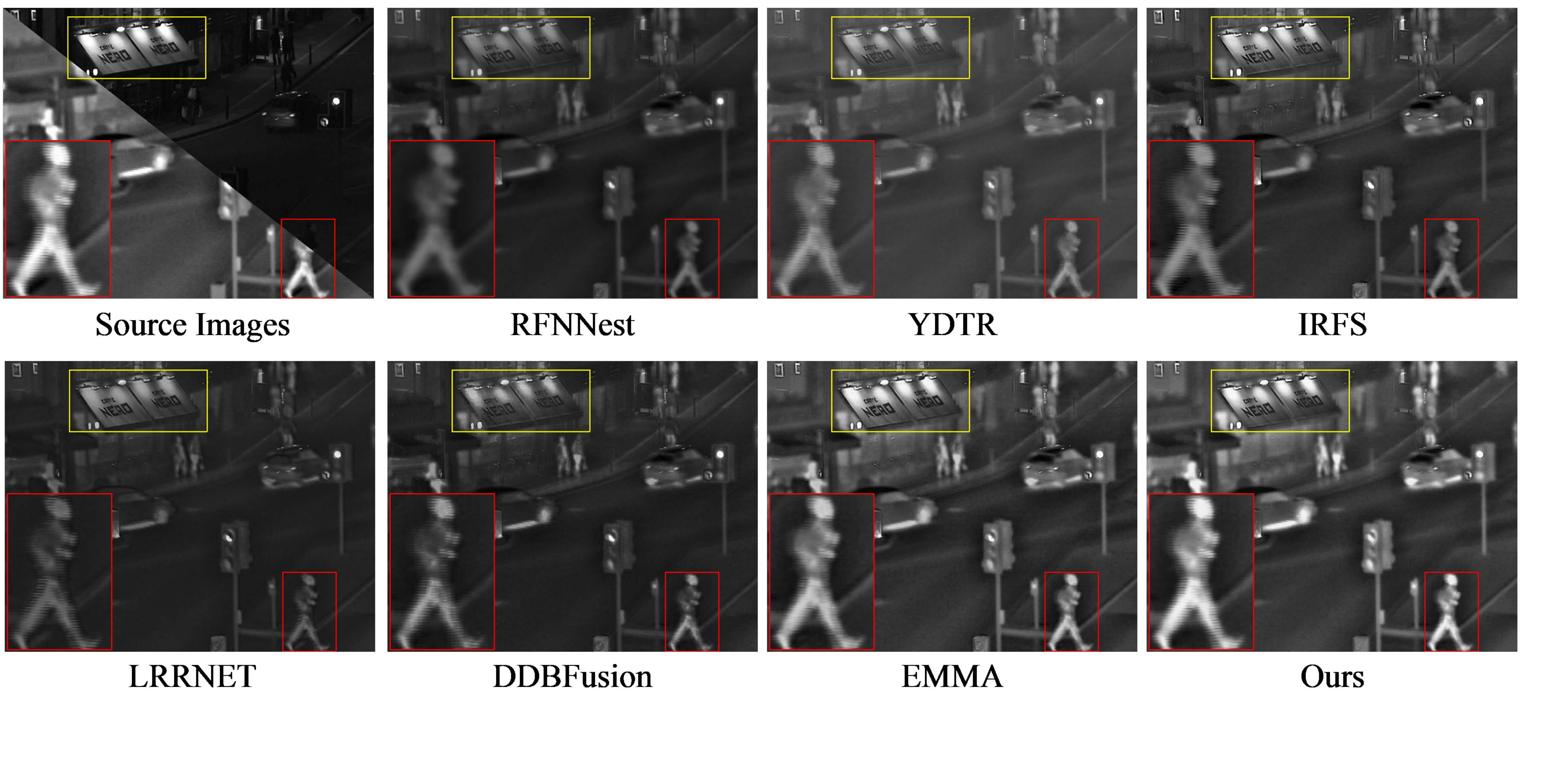}
	\caption{Qualitative comparison on the TNO dataset.}
	\label{FIG:7}
\end{figure}
\subsection{Experimental Details}
\label{subsec41}
All experiments are implemented using PyTorch on an NVIDIA Tesla GPU. During the training phase, we use $2,000$ pairs infrared-visible image pairs from the LLVIP dataset as the primary training set. Data augmentation techniques like random cropping and flipping are applied to increase the variety of training samples. Before being fed into the network, all input images were randomly cropped and resized to $256 \times 256$, followed by normalization to keep pixel values within the range of $[0,1]$. The parameters of the feature extraction module are initialized with the pre-trained weights of LRRNET \citep{rf13}. The batch size is configured to 6, and the network parameters are optimized using Adam with an initial learning rate of 0.0001. During the testing phase, we assess the model’s performance on the LLVIP, RoadScene, and TNO datasets. For quantitative analysis, we selected several commonly used evaluation metrics, including mutual information (MI), visual information fidelity (VIF), $Q_{abf}$ and structural similarity index measure (SSIM). Higher metric values demonstrate superior fusion performance.
\subsection{Comparison Experiments}
\label{subsec2}
We compare the proposed model with six image fusion methods, including RFNNest \citep{rf48}, YDTR \citep{rf20}, IRFS \citep{rf49}, LRRNET \citep{rf13}, DDBFusion \citep{rf50}, and EMMA \citep{rf51}. All the above methods are implemented using their official codes to ensure a fair comparison.
\subsubsection{Qualitative Comparison}
\label{subsubsec1}
Fig.~\ref{FIG:5}, ~\ref{FIG:6}, and ~\ref{FIG:7} show the qualitative results of different fusion methods on the LLVIP, RoadScene, and TNO datasets. From a visual perspective, our method demonstrates three significant advantages. First, it effectively highlights high-contrast and clear thermal targets. As shown in the first scene of Fig.~\ref{FIG:6} and in Fig.~\ref{FIG:7}, other methods often suffer from target information loss or attenuation when preserving thermal radiation salient targets, which affects target visibility and recognizability.  Second, our method maintains the rich texture and structural details during the fusion process. As shown in Fig.~\ref{FIG:5} and Fig.~\ref{FIG:6}, due to the high pixel intensity of infrared images, most fusion methods tend to cause color distortion and fail to retain sufficient texture and structural information. In contrast, our method excels in this aspect, successfully preserving details such as flower beds, pavement tiles, trees, and streetlights, while the overall color aligns better with human visual perception. Finally, our method effectively suppresses artifacts introduced by different modalities, such as thermal radiation noise and visible noise. As shown in the second scene of Fig.~\ref{FIG:6} and in Fig.~\ref{FIG:7}, compared to other methods, our fusion results exhibit significantly fewer artifacts in regions such as the sky and sunshades.  Additionally, It is important to remember that while our method slightly underperforms EMMA in preserving visible texture details on the LLVIP dataset, it significantly outperforms EMMA on the RoadScene and TNO datasets. This is mainly because EMMA’s training set consists primarily of low-light street scenes, which is consistent with the scene distribution of LLVIP, but differ significantly from RoadScene and TNO. This observation indicates that our method maintains stable fusion performance under different lighting conditions and scene categories, effectively mitigating the impact of dataset scene bias.

\subsubsection{Quantitative Comparison}
\label{subsubsec42}
Table ~\ref{Table1} presents the quantitative results on the LLVIP, RoadScene, and TNO datasets. Compared to other methods, our method performs better overall. The highest MI, VIF, and SSIM values demonstrate that our method better transfers rich content and structural information from the source images to the fused image, aligning more closely with human visual perception. Additionally, while our method achieves a slightly lower $Q_{abf}$ than EMMA on the LLVIP dataset, indicating slightly less texture detail preservation, it performs better on the RoadScene and TNO datasets. Particularly, our method is trained only on the LLVIP dataset without fine-tuning on other datasets, further validating its ability to mitigate the impact of dataset scene bias, thereby enhancing the model generalization ability.

\begin{table}
\caption{Quantitative comparison on three datasets (Bold: optimal, underline: suboptimal).}
\resizebox{\textwidth}{!}{
\begin{tabular}{c |c| c c c c c c c} 
 \hline
    Datasets&Metrics&RFNNest&YDTR&IRFS&LRRNET&DDBFusion&EMMA&Ours\\
 \hline
 
\multirow{4}{*}{LLVIP}&MI&2.5959&3.1878&2.5677&2.4245&2.4465&\underline{3.5676}&\textbf{3.7130}\\
                      &VIF&0.6452&0.6327&0.7433&0.5529&0.6698&\underline{0.8073}&\textbf{0.8417}\\
                &$Q_{abf}$&0.2956&0.3182&0.5700&0.4286&0.4282&\textbf{0.6435}&\underline{0.6103}\\
                &SSIM&0.3558&0.3617&0.4131&0.3208&0.3910&\underline{0.4265}&\textbf{0.4556}\\ \hline
\multirow{4}{*}{RoadScene}&MI&2.8759&2.9962&2.8542&3.0205&2.8450&\underline{3.3227}&\textbf{4.1864}\\
           &VIF&0.5174&0.5400&0.5465&0.5119&0.5454&\underline{0.5813}&\textbf{0.6875}\\
                &$Q_{abf}$&0.3278&0.4086&0.4086&0.3816&0.4210&\underline{0.4314}&\textbf{0.4739}\\
                &SSIM&0.4054&0.4590&\underline{0.4764}&0.3447&0.4605&0.4633&\textbf{0.5048}\\ \hline
\multirow{4}{*}{TNO}&MI&2.0351&2.4240&2.1023&2.5238&1.9793&\underline{2.9191}&\textbf{3.2799}\\
                      &VIF&0.5618&0.5462&0.5630&0.5508&0.5875&\underline{0.6428}&\textbf{0.6978}\\
                &$Q_{abf}$&0.3598&0.3657&0.4165&0.3696&0.3958&\underline{0.4836}&\textbf{0.4947}\\
                &SSIM&0.4056&0.4638&\underline{0.5034}&0.3971&0.4774&0.4862&\textbf{0.5120}\\
 \hline
\end{tabular}}
\label{Table1}
\end{table}

\begin{table}
\centering
\caption{Qualitative comparison of BAFFM ablation studies (Bold: optimal).}
\begin{tabular}{c |c| c c c c } 
 \hline
    Datasets&BAFFM&MI&VIF&$Q_{abf}$&SSIM\\
 \hline
 
\multirow{2}{*}{LLVIP}&w/o BAFFM&3.6569&0.7093&0.3206&0.3651\\
                      &Ours&\textbf{3.7130}&\textbf{0.8417}&\textbf{0.6103}&\textbf{0.4556}\\ \hline
\multirow{2}{*}{RoadScene}&w/o BAFFM&\textbf{4.2530}&0.6023&0.4285&0.4384\\
           &Ours&4.1864&\textbf{0.6875}&\textbf{0.4739}&\textbf{0.5048}\\ \hline
\multirow{2}{*}{TNO}&w/o BAFFM&3.0397&0.6457&0.4701&0.5046\\
                      &Ours&\textbf{3.2799}&\textbf{0.6978}&\textbf{0.4947}&\textbf{0.5120}\\
 \hline
\end{tabular}
\label{Table2}
\end{table}

\begin{figure}
        \centering
	\includegraphics[scale=0.2]{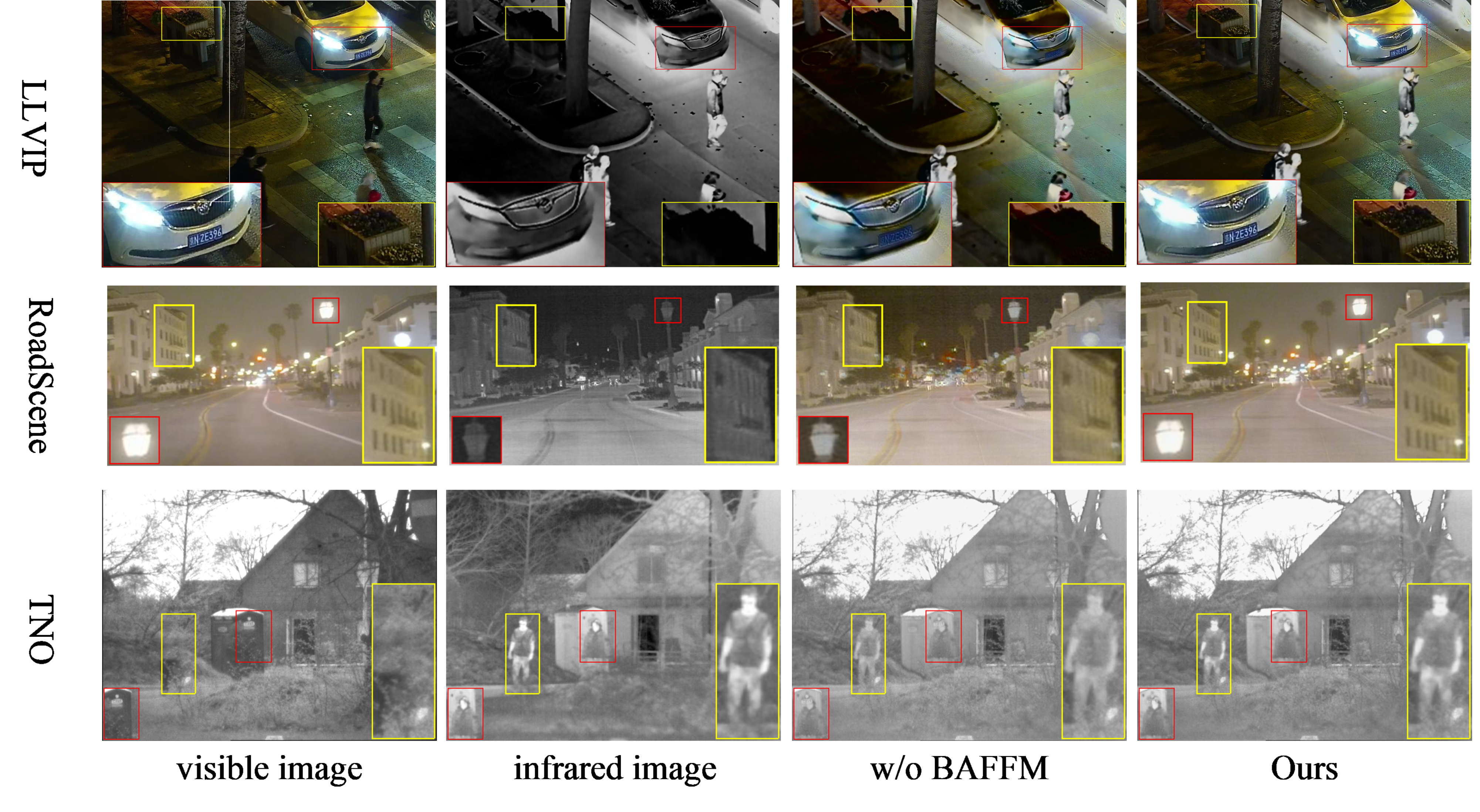}
	\caption{Qualitative comparison of BAFFM ablation studies}
	\label{FIG:8}
\end{figure}

\subsection{Ablation Studies}
\label{subsec43}
Rationality of BAFFM: To evaluate the effectiveness of the Back-door Adjustment based Feature Fusion Module (BAFFM) in the fusion process, we conducted ablation studies. Specifically, we removed the back-door adjustment mechanism and retained only the traditional feature fusion strategy to analyze the model fusion performance under different configurations. The qualitative and quantitative experimental results in Fig.~\ref{FIG:8} and Table ~\ref{Table2} demonstrate that BAFFM effectively eliminates the interference of dataset scene bias on model learning. This allows the model to more accurately assign fusion weights based on the complementarity of cross-modal information, significantly enhancing the fusion results quality.

Size of confounder dictionary: When stratifying the confounder $Z$, we carried out ablation studies on the dictionary size $N$. As presented in Table ~\ref{Table3}, the quantitative results show that when $N$ equals $25$ and $30$, the performance across multiple metrics remains similar, suggesting that further increasing the dictionary size provides minimal improvement. However, a larger $N$ requires storing more features and increases computational cost during optimization. Therefore, we ultimately set the confounder dictionary size $N$ to 25.

\begin{table}
\caption{Quantitative comparison of ablation studies on confounder dictionary size N (Bold: optimal, underline: suboptimal).}
\centering
\begin{tabular}{c |c| c c c c } 

 \hline
    Datasets&N&MI&VIF&$Q_{abf}$&SSIM\\
 \hline
 
\multirow{3}{*}{LLVIP}&20&3.5045&0.8129&\textbf{0.6238}&0.4500\\
                  &25&\underline{3.7130}&\textbf{0.8417}&\underline{0.6103}&\textbf{0.4556}\\
                  &30&\textbf{3.7238}&\underline{0.8374}&0.6049&\underline{0.4514}\\ \hline
\multirow{3}{*}{RoadScene}&20&3.7528&0.6169&0.4457&\textbf{0.5075}\\
           &25&\underline{4.1864}&\underline{0.6875}&\textbf{0.4739}&\underline{0.5048}\\ 
           &30&\textbf{4.3237}&\textbf{0.6946}&\underline{0.4714}&0.4972\\ \hline
\multirow{3}{*}{TNO}&20&2.9348&0.6513&0.4851&\textbf{0.5177}\\
                    &25&\underline{3.2799}&\textbf{0.6978}&\underline{0.4947}&\underline{0.5120}\\
                    &30&\textbf{3.4066}&\underline{0.6965}&\textbf{0.4952}&0.5111\\
 \hline
\end{tabular}
\label{Table3}
\end{table}

\section{Conclusion}
\label{sec5}
This paper proposes a novel causality-driven infrared and visible image fusion method. Specifically, we clarify the causalities between variables through tailored image fusion causal graph, providing a new theoretical framework and perspective for image fusion task. Building on this theoretical foundation, we further propose a Back-door Adjustment based Feature Fusion Module (BAFFM) to eliminate the confounding effects caused by dataset scene bias, ensuring fair participation of different scenes in the model training. Extensive experiments on standard infrared and visible image fusion datasets point out that the proposed method performs better than state-of-the-art alternatives.

\section*{Acknowledgments}
This work is supported by the Fundamental Research Program of Shanxi Province (No. 202303021211147), the
Special Program for Patent Transformation of Shanxi Intellectual Property Administration (No. 202302001), and
Shanxi Province Graduate Education Innovation Plan (No. 2024KY587).

\bibliographystyle{unsrt}







\end{document}